\documentclass{article}

\usepackage[preprint]{neurips_2026}

\usepackage{booktabs}
\usepackage{longtable}
\usepackage[utf8]{inputenc} 
\usepackage[T1]{fontenc}    
\usepackage{hyperref}       
\usepackage{url}            
\usepackage{booktabs}       
\usepackage{amsfonts}       
\usepackage{nicefrac}       
\usepackage{multirow}   
\usepackage{makecell} 
\usepackage{microtype}      
\usepackage{xcolor}         
\usepackage{graphicx}
\usepackage{amsmath}
\usepackage{amssymb}
\usepackage{tikz}
\definecolor{gold}{HTML}{BD820B}
\definecolor{silver}{HTML}{909090}
\definecolor{bronze}{HTML}{9A5F26}
\definecolor{softgreen}{RGB}{34,139,34}    
\definecolor{softred}{RGB}{220,20,60}      
\definecolor{softorange}{RGB}{255,140,0}   
\definecolor{softblue}{RGB}{30,144,255}    %
\newcommand*\circledd[1]{\tikz[baseline=(char.base)]{
            \node[shape=circle,draw,inner sep=0.15pt] (char) {#1};}}     
            
\newcommand{\first}[1]{%
    {#1\raisebox{0.8pt}{\footnotesize \color{gold} \circledd{1}}}%
}
\newcommand{\second}[1]{%
    {#1\raisebox{0.8pt}{\footnotesize \color{silver} \circledd{2}}}%
}

\setcitestyle{numbers,square}
\title{UGO: Unified Architecture for General Multi-Object Tracking by Segmentation}

\newcommand{\name}{{UGO}} 
\author{%
  Jer Pelhan, Alan Lukežič, Matej Kristan \\
  Faculty of Computer and Information Science, University of Ljubljana\\
  jer.pelhan@fri.uni-lj.si
}

\begin{document}

\maketitle

\begin{abstract}
General multi-object tracking (GMOT) tracks all instances of a user-specified category from a single first-frame exemplar. Prior work relies on bounding boxes and surrogate training, and struggles with non-rigid objects, crowded scenes, and distractors. We introduce UGO, a unified GMOT tracker that pairs a pretrained exemplar-conditioned detection head with an instance-propagation head in a common architecture. A novel training-free, energy-minimization consolidation method converts overlapping proposals into exclusive pixel-wise masks and detections, resolving over-segmentation, duplicates, and conflicts. A hierarchical memory spanning global and instance levels improves recall and per-instance segmentation accuracy using a new memory management protocol. UGO sets a new state-of-the-art on GMOT benchmarks and video object counting, and is competitive with specialist MOT methods, establishing a strong paradigm for unified, open-category multi-object tracking. The code will be available \href{https://github.com/jerpelhan/UGO}{here}.
\end{abstract}

\section{Introduction}
\label{sec:intro}

Given a single instance exemplar, e.g., specified by a bounding box in the first frame, the task of general multi-object tracking (GMOT) is to track \textit{all instances of the same category} throughout the video. 
This extends two classical problems: single-object tracking (SOT)~\cite{vot_tpami}, which extracts a trajectory of a single selected instance of an arbitrary category, and multi-object tracking (MOT)~\cite{dendorfer2021motchallenge}, which detects and associates multiple instances of a predefined class using a category-specific detector. 
GMOT thus unifies the generalization requirement of SOT with the coverage of MOT, while operating without category-specific training data, making it a fundamentally challenging problem.

\begin{figure}
  \centering
   \includegraphics[width=0.98\linewidth]{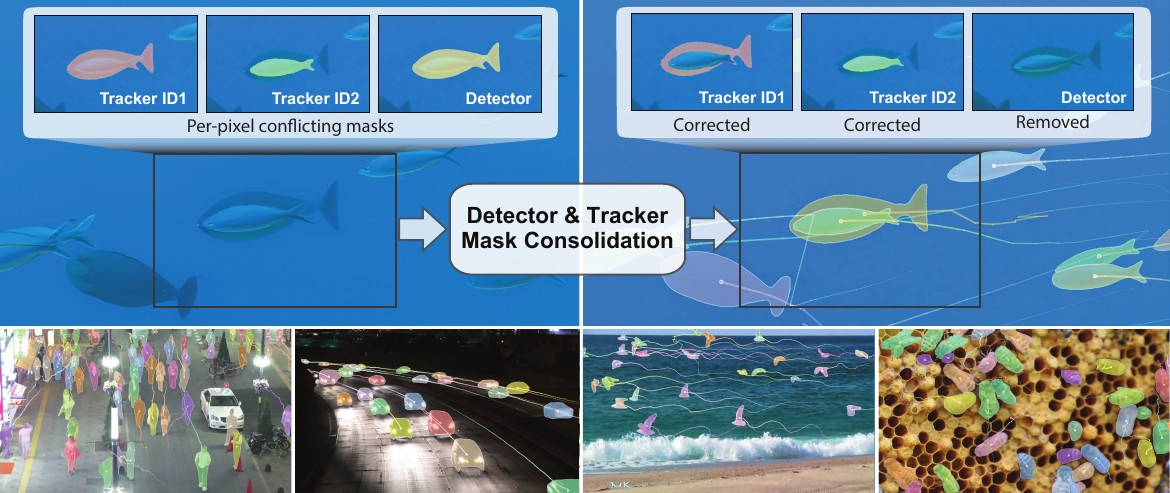}
    \caption{
    First row: Tracker ID=1 oversegments an occluding fish tracked by ID=2, while a single detector fires on both. \name{} resolves the oversegmentations and redundant detection by a novel consolidation module.
    Second row: The new paradigm enables robust general multi-object tracking by segmentation in cluttered dense scenes through occlusions.}
   \label{fig:firstfig}
\end{figure}

Most GMOT methods~\cite{huang2020globaltrack, wojke2017simple, mdp, famnet_mot} are direct adaptations of the classical online MOT paradigm, combining a detector, Kalman-based instance propagation, and Hungarian matching to associate detections with instance predictions. Alternatively, tracking and detection association is posed as an end-to-end trainable detection problem~\cite{siamese-detr} in the context of query-based detectors~\cite{detr}.
Since sufficiently diverse GMOT training data is unavailable, these methods rely on surrogate training on detection datasets, which cannot capture the complex dynamics of instance interactions.  Moreover, all current methods localize targets as bounding boxes, which may be adequate for approximating pedestrians, but poorly approximate general non-compact and deformable objects, making the association unreliable, particularly in crowded scenes with overlapping instances.

In general, all instances are better represented by segmentation masks, which enable pixel-level instance separation. 
This is explored in video object segmentation~\cite{AOT}, where masks of all tracked instances are jointly inferred. 
However, architectures for joint instance prediction do not scale to a large number of objects and require dedicated training sets.
Alternatively, recent video foundation models~\cite{sam2} hold the potential to independently track all instances, but these become unreliable under strong instance interactions and require external initialization at each new instance.

We address the aforementioned issues, by proposing UGO, a Unified General multiple Object tracker, that unifies recent advances in general object detection and the expressiveness of the video segmentation foundation models within a single architecture. 
UGO applies a video foundation backbone~\cite{sam2} with two lightweight heads, one for detecting instances corresponding to the user-provided exemplar, and another for frame-to-frame identity propagation. 
Both use the same segmentation module to predict calibrated per-pixel object presence beliefs, which may contain oversegmentation, double detections, and duplicated tracks (Figure~\ref{fig:firstfig}). 
These are resolved into unique, pixel-wise exclusive tracked instance segmentations and detections of new instances by a novel training-free {consolidation module}, formulated as an energy-minimization problem with a global exclusivity objective. 
The novel design avoids the need for extensive, diverse GMOT-specific training datasets to solve complex combinatorial optimization problems efficiently.
To ensure robustness to distractors and to align the detector with within-category instance appearance variation, we propose a {hierarchical adaptive memory} (HAM) composed of a detector category-level memory and individual per-instance memories. 
The memories are updated by jointly considering the tracks of all instances. UGO shows insensitivity to the hyperparameters commonly applied in MOT/GMOT settings, confirming its robust design.



Our contributions are: (i) a new general multi-object tracking paradigm with independently applied single-object trackers and detectors, coupled through mask consolidation and memory feedback to achieve a global coordination; (ii) a novel training-free energy-based consolidation module that delivers per-pixel exclusive segmentation masks from per-instance tracks and detections; (iii) a hierarchical adaptive memory (HAM) with a management protocol based on explicit failure detection, ensuring robust instance detection and identity propagation in the presence of distractors.
UGO unifies SOT, MOT, and GMOT in a single framework, achieving state-of-the-art performance and  defining a new paradigm in category-agnostic multi-object tracking.

\section{Related Work}

\noindent\textbf{Multi-Object Tracking (MOT).} In classical MOT pipelines such as SORT~\cite{bewley2016simple}, DeepSORT~\cite{wojke2017simple}, Tracktor~\cite{bergmann2019tracking}, and ByteTrack~\cite{zhang2022bytetrack}, detectors produce per-frame boxes or masks, which are then associated by motion or appearance cues. 
Mask-level variants~\cite{voigtlaender2019mots} (MOTS) extend this pipeline to segmentation. 
The methods excel in closed-world setups with strong class-specific detectors available, while their stability hinges on detector recall -- missed detections lead to trajectory fragmentation -- thus they underperform on long-tail categories. Recent works such as TrackFormer~\cite{meinhardt2022trackformer} and TransTrack~\cite{sun2020transtrack} use DETR-like queries for joint detection and identity propagation. However, as noted in~\cite{ Gao_2025_CVPR}, this creates a conflict between category semantics (detection) and instance specificity (association). In addition, most methods require per-category training and lack pixel-level exclusivity across similar instances, causing overlaps and ID switches in crowded scenes.

\noindent\textbf{General single object trackers (SOT).}
Modern SOT enable tracking of any instance and are highly robust to appearance changes and occlusion~\cite{sam2,dam4sam,cutie}, while recent memory mechanisms improve reliability under distractors~\cite{dam4sam,vots2025}. 
However, SOT lacks scene-level competition and cannot ensure mutual exclusivity among many look-alikes; naively running multiple SOT instances on MOT datasets is prone to identity swaps. 
But SOT methods cannot discover new instances appearing in the video, and optimize per-target propagation rather than frame-level joint reasoning over all instances. 

\noindent\textbf{General \& open-category tracking.}
General multi-object tracking (GMOT) extends MOT to tracking any instance of an unseen category, specified by a single instance exemplar.
The GMOT-40 benchmark~\cite{bai2021gmot} established a standardized evaluation setting, demonstrating that a straightforward combination of a one-shot detector~\cite{huang2020globaltrack} and a target association module~\cite{famnet, mdp, bochinski2017high} provides a simple, yet limited baseline.
Subsequent methods, including PLGMOT~\cite{liu2024prototype} and S-DETR~\cite{siamese-detr}, introduce exemplar-conditioned detection and transformer-based association. While PLGMOT enables instance discovery, its tracking-by-detection design limits recall. S-DETR improves this via exemplar-conditioned queries and dynamic track reuse. However, both remain restricted to box-level reasoning and NMS-based conflict resolution, causing identity switches in crowded scenes. Recently, SAM3~\cite{sam3} extended single-object tracking toward open-vocabulary tracking, integrating detection capabilities, but it still lacks explicit pixel-level interaction reasoning among tracked instances.

\section{A Unified General Multiple Object Tracker}  \label{sec:method}

\begin{figure*}[h!]
  \centering
   \includegraphics[width=\linewidth]{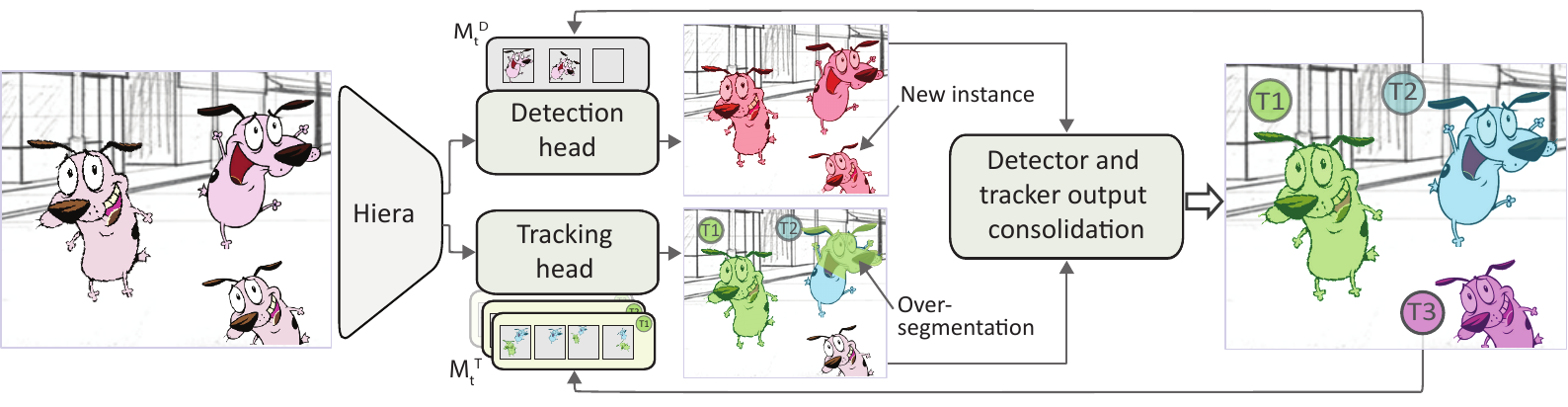}
   \caption{
   UGO unifies detection and frame-to-frame association in a single network. Detector and instance tracker heads produce compatible per-pixel outputs, which enables optimization-based, training-free consolidation and robust tracking.
   }
   \label{fig:arch_overview}
\end{figure*}

Given a sequence of $N$ video frames $\{ \mathbf{I}_t \}_{t=1:N}$ and an instance category exemplar bounding box $\mathbf{b}^{E}_{1}$ provided in the first frame, the task is to track all instances of the same category in the video. We propose a unified general multiple object tracker (\name{}) (Figure~\ref{fig:arch_overview}).
At time-step $t$, \name{} already tracks some of the instances from the previous frame and proceeds in the current frame as follows. The input image $\mathbf{I}_t \in \mathbb{R}^{H \times W \times 3}$ is encoded by the Hiera~\cite{hiera} backbone into $\mathbf{F}_t \in \mathbb{R}^{h \times w \times c}$, with spatial resolution $h \times w$ and $c$ feature channels. The features are passed to two light-weight heads. The first is the SAM2~\cite{sam2} video segmentation head that propagates tracked instances from the previous time-step (Section~\ref{sec:method_tracker}). 
The second head is based on GECO2~\cite{geco2} and detects all new instances and potentially also those already tracked, matching the exemplar category (Section~\ref{sec:method_detector}).
Both heads apply a pretrained SAM2 mask decoder, delivering compatible per-pixel logits, which enables consolidation of detector and tracker outputs into mutually exclusive masks (Section~\ref{sec:method_consolidation}).
The consolidation influences memory updating: the resolved conflicts trigger the affected tracker memory updates. The individual trackers are thus implicitly coupled through shared pixel-level decisions, forming a feedback loop that reduces future overlaps and stabilizes identity tracking (Section~\ref{sec:method_memory}).
This substantially simplifies trajectory management (Section~\ref{sec:method_lifecycle_management}).

\subsection{Frame-to-frame instance propagation}  \label{sec:method_tracker}

A visual model $\mathcal{M}^{T}_{t-1,k}$ is maintained for each $k$-th tracked instance in form of a memory (Section~\ref{sec:method_memory}).
For each instance $k$, SAM2 head attends the backbone features to the corresponding memory and produces a logit map $\mathbf{T}_{t,k} \in \mathbb{R}^{H\times W}$, which reflects a belief that each pixel contains the instance, conditioned on its visual model. 
The output of the frame-to-frame propagation module is thus a set of per-instance tracker logits $\mathcal{T}_t = \{ \mathbf{T}_{t,k} \}_{k=1:N_\mathcal{T}}$.

\subsection{General instance localization}  \label{sec:method_detector}

Since only a single user-provided exemplar of the tracked instance category is available in the first frame, all other instances (in the first frame and also entering in later frames) have to be detected to initialize new per-instance trackers. 
We utilize the general-object detector GECO2~\cite{geco2}, a light-weight detection head, operating on the SAM2 backbone. 
The object category is specified by exemplar memory $\mathcal{M}_t^{D} = \{ \mathbf{b}^E_i \}_{i=1:N_B}$, which is initialized by the first-frame exemplar $\mathbf{b}^{E}_{1}$ and is updated during tracking (Section~\ref{sec:method_memory}). 
The detection head localizes the instances and applies the SAM2 segmentation head to produce detection logits 
$\mathcal{D}_t = \{ \mathbf{D}_{t,i} \}_{i=1:N_{\mathcal{D}}}$,
where $\mathbf{D}_{t,i} \in \mathbb{R}^{H\times W}$ reflects the belief of each pixel belonging to the instance, conditioned on the memory. 






\subsection{Detector and tracker output consolidation}
\label{sec:method_consolidation}

The use of the same pre-trained SAM2 segmentation head by the detector and tracker ensures calibrated and compatible logits in $\mathcal{T}_t$ and $\mathcal{D}_t$, which could be na\"ively converted to masks by thresholding at zero.
However, these masks indicate all pixels that \textit{may} belong to the detector/tracker instance, not accounting for other instances. 
Thus, the masks will not be mutually exclusive, with some objects covered by both detectors and trackers, multiple trackers including distractor objects, and masks of different objects potentially overlapping due to over-segmentation  (Figure~\ref{fig:drm_update}).
Because of these nontrivial interactions, the classical MOT bipartite~\cite{hun_matching,NeuralSolver} or quadratic~\cite{wright2015} optimization, even if redesigned to operate at the per-mask level, and disregarding the significant compute complexity, cannot be applied for conflict resolving.

To address these issues, we propose a new formulation that ensures mutually exclusive masks. 
We start by defining the logits tensor $\textbf{L} \in \mathbb{R}^{H \times W \times S}$ obtained by concatenating the $N_\mathcal{D}$ detector and $N_\mathcal{T}$ tracker logit maps with a small constant background logit map, i.e., $\mathbf{L} = \mathrm{cat}(\mathcal{D}_t, \mathcal{T}_t, \mathbf{1} \lambda_\mathrm{BG})$. 
For notation compactness, let $\mathbf{L}(\mathbf{x},s)$ retrieve the logit value of map $s$ at pixel position $\mathbf{x}$. Next, let $\mathbf{Y} \in [1,\dots,S]^{W \times H}$ denote the mutually-exclusive pixel labeling with $\mathbf{Y}(\mathbf{x})$ retrieving the logit map identity assigned to pixel $\mathbf{x}$. 

The solution for $\mathbf{Y}$ should ensure 
(i) a unique label on each pixel, 
(ii) among the masks competing for the same pixels, the trackers should be preferred over detectors, and 
(iii) among the competing tracker masks, the trackers with masks supported by detections should be preferred. 
Such labeling can be obtained by minimizing the objective
\begin{equation} 
    \label{eq:loss1}
        \hat{\mathcal{L}}(\mathbf{Y}) = 
        -\sum_{\mathbf{x}\in \Omega \mathrm{ ; } s = \mathbf{Y}(\mathbf{x})} \log\Big( \mathbf{L}(\mathbf{x},s) \cdot \Theta_1(s) \cdot \Theta_2(s) \Big),
\end{equation} 
where $\Omega \!=\! \{1,\dots,H\}\!\times\!\{1,\dots,W\}$ denotes the set of pixel positions, and $\Theta_1(\cdot)$ and $\Theta_2(\cdot)$ are unitary potentials enforcing the required properties (i-iii) of the optimal labeling. The first potential $\Theta_1(\cdot)$ is defined as 
\begin{equation}
    \label{eq:theta1}
       \Theta_1(s) = 
            \begin{cases}
                    \frac{1}{2}\theta_0 & \text{; } s \leq N_\mathcal{D}\\
                    \theta_0 + \underset{d \in \mathcal{D}_t}{\max} \mathrm{IoU} \big( \mathbf{Y}\equiv s,\, d \big) & \text{; } N_\mathcal{D} < s \leq N_\mathcal{DT} \\
                    1 & \text{; } s \equiv S 
            \end{cases}
\end{equation}
where $N_\mathcal{DT}=N_\mathcal{D}+N_\mathcal{T}$, 
$\theta_0$ is a constant enforcing preference of trackers over detections ($\theta_0$ vs $\frac{1}{2}\theta_0$), and the IoU term prefers trackers better supported by the detections, assuming a labeling $\mathbf{Y}$. 
The second potential $\Theta_2(\cdot)$ reduces the scores of instances whose masks induced by labeling $\mathbf{Y}$ significantly deviate from their initially computed masks, indicating the other instances have absorbed their pixels, 
\begin{equation}
    \label{eq:theta2}
    \Theta_2(s) = 
    \begin{cases}
        \mathrm{IoU}(\mathbf{Y}\equiv s, L(:,s)>0) & \text{; } s \leq N_\mathcal{DT} \\
        1 & \text{; } s \equiv S
    \end{cases}.
\end{equation}

Since (\ref{eq:loss1}) is not separable in $\mathbf{Y}(\mathbf{x})$, we introduce auxiliary variables $\mathbf{A}_1 \in \mathbb{R}^{1 \times S}$ and $\mathbf{A}_2 \in \mathbb{R}^{1 \times S}$, leading to a surrogate objective

\begin{equation}
\label{eq:loss2}
\mathcal{L}(\mathbf{Y}) =
-\sum_{\mathbf{x}\in\Omega,\; s=\mathbf{Y}(\mathbf{x})}
\log\!\left(\mathbf{L}(\mathbf{x},s)\mathbf{A}_1(s)\mathbf{A}_2(s)\right)
+ \sum_{s=1}^{S} N_s
\left[
\log^2\!\frac{\mathbf{A}_1(s)}{\Theta_1(s)} +
\log^2\!\frac{\mathbf{A}_2(s)}{\Theta_2(s)}
\right],
\end{equation}

with $N_s$ the number of pixels assigned to label $s$, 
which can be optimized by MM-based~\cite{Lange2016MM} iterations of the following steps (see supplementary material for full derivation). 

Assuming a constant $\mathbf{Y}^{(k-1)}$, estimated at previous step, taking derivatives of (\ref{eq:loss2}) w.r.t. $\log\mathbf{A}_1$ and $\log\mathbf{A}_2$, and equating to zero, leads to the update,
\begin{equation} 
    \label{eq:step1}
        \mathbf{A}_1^{(k)}(s) \propto \Theta_{1|\mathbf{Y}^{(k-1)}}(s) \textrm{ ; } 
        \mathbf{A}_2^{(k)}(s) \propto \Theta_{2|\mathbf{Y}^{(k-1)}}(s).
\end{equation} 
Keeping $\mathbf{A}_1$ and $\mathbf{A}_2$ constant, the cost is separable across pixels, leading to maximization over $\mathbf{Y}$ as
\begin{equation} 
    \label{eq:step2}
        \mathbf{Y}^{(k)}(\mathbf{x}) = \operatorname*{argmax}_{s \in \{1:S\}} \big( 
        \mathbf{L}(\mathbf{x},s) \cdot 
        \mathbf{A}_1^{(k)}(s) \cdot \mathbf{A}_2^{(k)}(s) 
        \big). 
\end{equation} 
The optimization thus gradually re-assigns pixels claimed by several detector/tracker instances to either of them, and if a particular instance is poorly supported by the remaining pixels, it is automatically turned off. 
The labels $\mathbf{Y}^{(0)}$ are initialized by per-pixel argmax over the logits $\mathbf{L}$, while the optimization converges within a few iterations. 

Finally, severely reduced masks are removed: any label $s$ in $\mathbf{Y}$, whose mask IoU with the initialization mask is smaller than a fixed threshold, 
is reassigned to the background, i.e., $s=S$. 
\name~uses a permissive threshold for trackers $\tau_\textrm{lo}=0.2$ (since small overlap may be due to distractor correction), and, to ensure a high recall, we set the threshold for detector masks to a moderate level $\tau_\mathrm{md}=0.6$ (ablated in supplementary material).




\subsection{Instance track lifecycle management}  \label{sec:method_lifecycle_management}

After consolidation, only two types of mutually-exclusive masks remain: (i) masks corresponding to instance trackers and (ii) masks corresponding to newly detected instances.
Standard MOT rules are applied next for trajectory management. 
\\
\noindent\textbf{Initialization \& Update.} 
New single-target trackers are initialized on detection masks, while the instance trackers are updated with their consolidated masks.
\\
\noindent\textbf{Termination.} A tracker is flagged for termination when its mask collapses due to drift, or when the object leaves the field of view. In practice, the flag is raised when consolidation returns an empty mask.
If the condition is met for ten consecutive frames, the tracker is terminated.
\\
\noindent\textbf{Initialized track validation.}
To remove tracks initialized on a false positive detection, they are classified as valid only after frequently confirmed by the detector.
As a standard rule, tracks with 30$\%$ of their masks confirmed are declared valid, where a tracker mask $M_{t,k}$ is considered confirmed if its IoU with any of the detectors is greater than the moderate $\tau_{\text{md}}$, i.e.,
\begin{equation}
    \label{eq:support}
    \max_{d \in \mathcal{D}_t} \operatorname{IoU}(M_{t,k}, d>0) 
    \ge \tau_{\text{md}}.
\end{equation}



\subsection{Hierarchical Adaptive Memory} 
\label{sec:method_memory}
UGO employs a hierarchical adaptive memory (HAM) to represent the targets at two levels of detail. 
The category-level memory is used by the detector, while per-instance-level memory is used by the tracker. 
The two are continually updated to improve detection capabilities and to ensure instance-level tracking robustness. Figure~\ref{fig:mem_update} overviews HAM.

\textbf{Category-level memory} 
$\mathcal{M}_t^{D} = \{ \mathbf{b}^E_i \}_{i=1:N_B}$, 
always contains the user-provided exemplar $\mathbf{b}_1^E$ and holds a FIFO buffer with three slots, updated by the exemplars proposed from reliable per-instance trackers. 
At time-step $t$, we consider all trackers with at least ten frames long trajectories, whose current mask $M_t$ highly overlaps with a detector, i.e.,
$\operatorname{IoU}(M_t, d>0) \ge \tau_{\text{hi}}$, where $\tau_{\text{hi}} = 0.9$.
Among these, the current mask-fitted-bounding-box from the most frequently confirmed (\ref{eq:support}) trajectory updates the category-level memory $\mathcal{M}_t^{D}$.



\begin{figure}
  \centering
   \includegraphics[width=\linewidth]{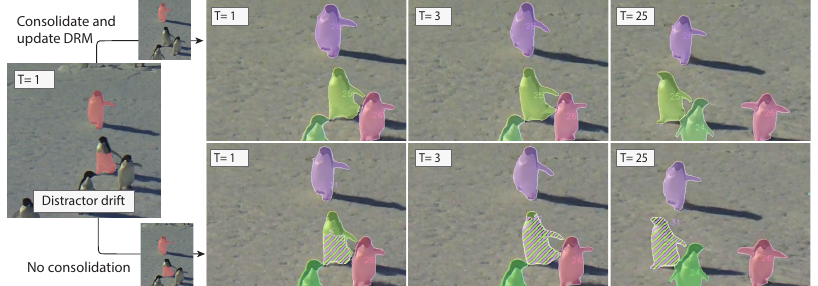}
   \caption{
   Oversegmented object at $T=1$ and tracking in subsequent frames with and without consolidation-driven DRM update. Individual colors indicate an ID, with dashed pattern indicating overlapping IDs.
   Consolidation-based memory update leads to distractor resolution, while ignoring it leads to tracker failure. 
   }
   \label{fig:drm_update}
\end{figure}

{\bf Instance-level memory} 
$\mathcal{M}_{t,k}^{T}$ stores segmented examples of $k$-th instance and is used for its localization by the tracker. 
As in~\cite{dam4sam}, the memory is split into distractor-resolving memory (DRM), responsible for discriminating between the target and visually-similar objects, and recent appearance memory (RAM), responsible for frame-to-frame segmentation accuracy -- each a FIFO buffer with three DRM and four RAM slots. 
Both are updated from non-empty tracker outputs, with RAM updated as in~\cite{dam4sam}, while we introduce a new DRM update protocol.
The standard assumption~\cite{dam4sam} that the initialization frame provides an iconic target with ground-truth segmentation does not hold in GMOT, where trackers are initialized from potentially inaccurate detections. Therefore, the initialization frame should not be retained in DRM indefinitely~\cite{sam2,dam4sam}. 
We instead leverage detector–tracker consolidation to identify frames with distractors. DRM is updated when the tracker mask before and after consolidation no longer reflect a high agreement (IoU$<\tau_{\text{hi}}$), while the consolidated mask is detector-confirmed (Eq.~\ref{eq:support}), indicating correction of distractor over-segmentation. Figure~\ref{fig:drm_update} visualizes the benefits of using the proposed DRM updating scheme.


\section{Experiments}

We follow the standard GMOT evaluation protocol~\cite{bai2021gmot}, where a single target exemplar is provided in the first frame.
Performance is evaluated by MOT measures, with the primary being MOTA, which integrates false positives, false negatives, and identity switches. 
Where possible, we also report HOTA~\cite{luiten2021hota}, which jointly measures detection (DetA)  and association (AssA) accuracy and provides a more reliable overall assessment than MOTA~\cite{luiten2021hota}.
Several auxiliary measures are used. Identity preservation is measured by IDP, IDR, and their harmonic mean IDF1. Trajectory quality is quantified by MT, PT, and ML, denoting the number of mostly tracked (\(>80\%\)), partially tracked (\(20\text{-}80\%\)), or mostly lost (\(<20\%\)) identities, respectively. We also report false positives (FP), false negatives (FN), F1 score, identity switches (IDSw), and fragmentations (FM).

\textbf{Implementation details.} 
We use the pretrained SAM2.1 tracking module~\cite{sam2} with a pretrained GeCo2~\cite{geco2} head, both operating on the same SAM2.1 Hiera-L backbone. The consolidation optimization (Section~\ref{sec:method_consolidation}) runs for $N_\text{iter}=3$ iterations with parameter $\theta_0=0.5$. The same thresholds for low, moderate and high overlaps ($\tau_\mathrm{lo}=0.2$, $\tau_\mathrm{md}=0.6$ and $\tau_\mathrm{hi}=0.8$) are used in all experiments.
\name{}
tracks 100 objects at $\sim2.1$ FPS on a single A100 GPU.

\subsection{Comparison with general multi-object trackers}
\label{sec:generalist_MOT}

\name{} is compared on GMOT-40~\cite{bai2021gmot} benchmark with GMOT trackers, including the five versions of the current best state-of-the-art S-DETR~\cite{siamese-detr} (Table~\ref{table:gmot}). 
\name{} consistently outperforms all competitors. 
It outperforms S-DETR~\cite{siamese-detr} by $23\%$ MOTA, reflecting joint improvements in detection quality, association accuracy, and long-term trajectory consistency.
It also outperforms S-DETR~\cite{siamese-detr} by 52\% IDF1, indicating a substantially better instance identity preservation.
We also evaluate SAM3~\cite{sam3} with default parameters (see supplementary), which only supports prompt-based tracking -- we thus initialize it with the category names of target objects.
Despite this semantic advantage, \name{} surpasses SAM3 by 6\% in HOTA and 22\% in MOTA, demonstrating markedly stronger instance coverage and overall tracking robustness.

 \begin{table*}
\setlength{\tabcolsep}{4pt}
\caption{State-of-the-art comparison on the GMOT-40~\cite{bai2021gmot} benchmark.}
\label{table:gmot}
\centering
\resizebox{\textwidth}{!}{
\begin{tabular}{l|llccccccccc}
\toprule
& Detector & Tracker & IDF1$\uparrow$ & MT$\uparrow$ & ML$\downarrow$ & FP$\downarrow$ & FN$\downarrow$ & F1$\uparrow$ & IDSw$\downarrow$ & MOTA$\uparrow$ & HOTA$\uparrow$ \\
\hline


\multirow{9}*{\rotatebox{90}{Open-Vocabulary}} 
& \multirow{4}*{OVTrack~\cite{li2023ovtrack}}
& DeepSORT~\cite{wojke2017simple} & 21.2 & 165 & 1367 & 49984 & 160378 & 47.7 & 1470 & 20.2 & - \\
&& ByteTrack~\cite{zhang2022bytetrack} & 20.6 & 164 & 1345 & 51356 & 156329 & 49.1 & 1669 & 19.9 & - \\
&& BoT-SORT~\cite{aharon2022bot} & 20.3 & 167 & 1328 & 45721 & 163378 & 47.1 & 3278 & 20.0 & - \\
&& TbQ~\cite{siamese-detr}-SwT & 18.7 & 186 & 1304 & 50784 & 154893 & 49.7 & 6381 & 21.3 & - \\
\cline{2-12}

& \multirow{4}*{GLIP-T~\cite{li2022grounded}}
& DeepSORT~\cite{wojke2017simple} & 41.6 & 401 & 877 & 46610 & 141330 & 55.0 & 2892 & 25.5 & - \\
&& ByteTrack~\cite{zhang2022bytetrack} & 45.1 & 447 & 746 & 52591 & 131759 & 57.5 & 2706 & 27.0 & - \\
&& BoT-SORT~\cite{aharon2022bot} & 49.1 & 553 & 643 & 51308 & 133462 & 57.1 & 4675 & 27.3 & - \\
&& TbQ~\cite{siamese-detr}-SwT & 39.8 & 581 & 592 & 49470 & 136602 & 56.3 & 9972 & 27.5 & - \\

\cline{2-12}

& SAM3~\cite{sam3}&  & 72.8  & 1209 & 276 & 70994 & 56131 & 75.9 & \textbf{990} & 50.2 & 60.1 \\
\hline
\multirow{12}*{\rotatebox{90}{Template-Based}} 
& \multirow{4}*{GTrack~\cite{huang2020globaltrack}}
& DeepSORT~\cite{wojke2017simple} & 24.4 & 72 & 1363 & \textbf{9000} & 208818 & 30.4 & 1315 & 14.5 & - \\
&& ByteTrack~\cite{zhang2022bytetrack} & 32.1 & 178 & 1069 & 23881 & 181829 & 42.0 & 1791 & 19.1 & - \\
&& BoT-SORT~\cite{aharon2022bot} & 34.0 & 251 & 978 & 22229 & 176991 & 44.3 & 7375 & 19.4 & - \\
&& TbQ~\cite{siamese-detr}-SwT & 27.4 & 213 & 1066 & 13507 & 182376 & 43.0 & 6407 & 20.6 & - \\
\cline{2-12}

& \multirow{4}*{S-DETR~\cite{siamese-detr}}
& DeepSORT~\cite{wojke2017simple} & 41.8 & 382 & 773 & 47336 & 124257 & 60.6 & 5131 & 31.1 & - \\
&& ByteTrack~\cite{zhang2022bytetrack} & 41.4 & 331 & 764 & 53417 & 104765 & 65.7 & 4204 & 33.7 & - \\
&& BoT-SORT~\cite{aharon2022bot} & 47.5 & 431 & 674 & 45769 & 119288 & 62.4 & 6775 & 34.1 & - \\
&& TbQ~\cite{siamese-detr}-SwT & 42.8 & 504 & 666 & 44882 & 107894 & 66.0 & 11664 & 35.9 & - \\
&& TbQ~\cite{siamese-detr}-SwB & 51.3 & 1083 & 278 & 44390 & 68189 & 77.0 & 11252 & 50.0 & - \\

\cline{2-12}
& {\name{}}
&& \textbf{77.09} & \textbf{1349} & \textbf{170} & 58165 & \textbf{39670} & \textbf{81.3} & 1136 & \textbf{61.4} & \textbf{63.8} \\
\bottomrule
\end{tabular}}
\end{table*}

\begin{figure}
  \centering
   \includegraphics[width=\linewidth]{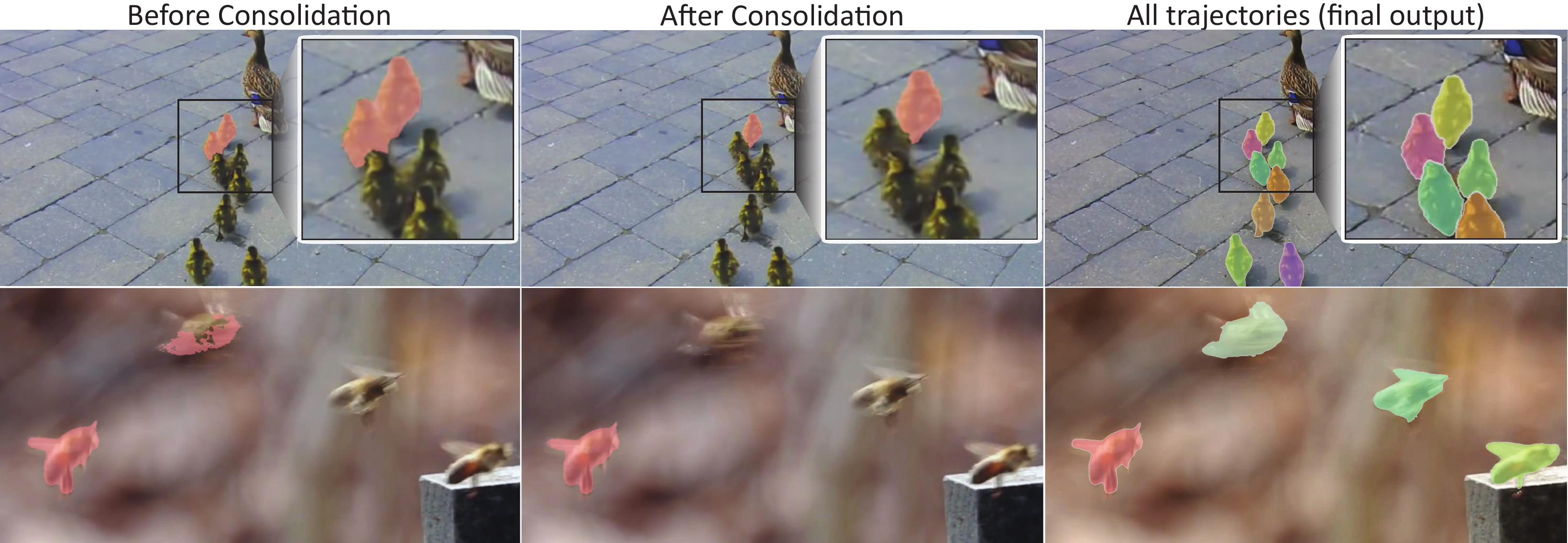}
   \caption{
   Incorrect masks spreading over several objects are corrected by the proposed detector-tracker consolidation method, enabling accurate memory updates and preventing drifts. 
   }
   \label{fig:mask_optimization}
\end{figure}

\name{} achieves the highest \textit{mostly tracked rate} (MT), surpassing S-DETR~\cite{siamese-detr} by 25\% and the lowest (-39\%) \textit{mostly lost rate} (ML), demonstrating a superior tracking stability. 
This highlights the effectiveness of the proposed pixel-level consolidation and evidence-gated memory mechanisms in enforcing consistent, globally coordinated identity assignment.
As illustrated in Figure~\ref{fig:mask_optimization}, the proposed consolidation resolves overlapping masks into mutually exclusive, pixel-consistent instance segmentations. By enforcing competition at the pixel level rather than relying on box-level matching (standard in prior SOTA), consolidation resolves ambiguous shared regions and corrects over-segmentation errors before they accumulate temporally. This prevents identity switches and trajectory fragmentations, especially in dense scenes with visually similar instances.

Compared to S-DETR, \name{} reduces false negatives by 42\% at a cost of 31\% higher FP rate, while it overall produces a better temporal coverage of target-category instances, as observed in 6\% higher F1 score.
Inspection of S-DETR code revealed usage of sequence-specific thresholds to reduce false positives, while \name{} relies on a fixed setting, indicating its gains come from architecture. Overall, \name{} achieves state-of-the-art performance with strong cross-category generalization.

\subsection{Comparison with specialist trackers}
\label{sec:specialist_MOT}

Next, we compare \name{} on benchmarks with trackers specialized for individual object categories. 
These setups are particularly challenging for generalist \name{}, since it has not been fine-tuned per category, unlike its competitors.

\textbf{Evaluation on AnimalTrack.}
AnimalTrack benchmark~\cite{animaltrack} evaluates classical MOT trackers on 10 different animal species. The benchmark provides training sets with category-specific labels to train the MOT trackers for each test category.
Since iconic exemplars are not provided, \name{} randomly selects five ground-truth bounding boxes from the first frame to define the target category.

Table~\ref{tab:animal} shows that \name{} outperforms all state-of-the-art methods. 
\name{} surpasses the strongest specialist method QDTrack~\cite{qdtrack} by 26\% HOTA and 27\% IDF1. 
Remarkably, \name{} tracks $2\times$ more instances than competing trackers, while producing $4\times$ fewer identity switches and half the number of fragmentations.
We additionally evaluate SAM3~\cite{sam3}, following Sec.~\ref{sec:generalist_MOT}. 
Since SAM3 relies on text prompts rather than exemplar conditioning, it benefits from an explicit semantic specification of the target (e.g., ``geese''), whereas \name{} must infer the category purely from visual evidence in a few exemplars.
Despite its semantic advantage and large-scale detection training, \name{} surpasses SAM3 by 14\% HOTA and 23\% MOTA, demonstrating superior performance.

\begin{table*}
\centering
\caption{State-of-the-art comparison on AnimalTrack~\cite{animaltrack}.}
\resizebox{\textwidth}{!}{
\setlength{\tabcolsep}{5.5pt}
\begin{tabular}{lccccccccccc}
\toprule
Method & HOTA$\uparrow$ & MOTA$\uparrow$ & IDF1$\uparrow$ & IDP$\uparrow$ & IDR$\uparrow$ & MT$\uparrow$ & PT & ML$\downarrow$ & F1$\uparrow$ & IDSw$\downarrow$ & FM$\downarrow$ \\
\midrule






ByteTrack~\cite{zhang2022bytetrack} & 40.1 & 38.5 & 51.2 & 64.9 & 42.3 & 310 & 465 & 329 & 0.631 & 1309 & 3513 \\

IOUTrack~\cite{bochinski2017high} & 41.6 & \textbf{55.7} & 45.7 & 51.9 & 40.7 & 388 & 454 & 262 & 0.762 & 4639 & 5259 \\

SORT~\cite{bewley2016simple} & 42.8 & 55.6 & 49.2 & 58.5 & 42.4 & 333 & 470 & 301 & 0.749 & 2530 & 3730 \\

OMC~\cite{liang2022one} & 43.0 & 53.4 & 50.3 & 61.8 & 42.4 & 324 & 478 & 302 & 0.735 & 4938 & 7162 \\

Tracktor++~\cite{bergmann2019tracking} & 44.2 & 55.2 & 51.0 & 58.5 & 45.1 & 364 & 472 & 268 & 0.751 & 1976 & 4149 \\

TransTrack~\cite{sun2020transtrack} & 45.4 & 48.3 & 53.4 & 63.4 & 46.1 & 327 & 416 & 361 & 0.705 & 1978 & 6459 \\

QDTrack~\cite{qdtrack} & 47.0 & \textbf{55.7} & 56.3 & 65.6 & 49.3 & 367 & 420 & 317 & 0.752 & 1970 & 5656 \\

SAM3~\cite{sam3} & 52.1 & 43.9 & 66.4 & \textbf{68.4} & 64.4 & 642 & 280 & 188 & 0.713 & 470 & 2588 \\

\name{} & \textbf{59.4} & 53.8 & \textbf{71.7} & 67.6 & \textbf{76.3} & \textbf{700} & 254 & \textbf{156} & \textbf{0.783} & \textbf{463} & \textbf{2490} \\

\bottomrule
\end{tabular}}
\label{tab:animal}%
\end{table*}

The primary limitation of \name{} is a higher FP rate, resulting in a 3\% lower MOTA than QDTrack~\cite{qdtrack}. 
Inspection reveals that a third of FPs originate from the \textit{goose\_3} sequence, where similar bird species are also detected, reflecting ambiguity in exemplar-based category specification.
Nevertheless, \name{} achieves superior identity quality and tracking performance, highlighting strong generalization without category-specific training.

\textbf{Evaluation on MOT17.}
We next evaluate \name{} on the human-centric MOT17 benchmark~\cite{MOT16}, again with the same setup as in AnimalTrack.
Table~\ref{tab:mot17} reports a comparison with three standard MOT baselines, and state-of-the-art OC-SORT~\cite{OC-SORT}.
\name{} outperforms all MOT baseline specialists in HOTA, demonstrating strong joint detection-association reasoning, but lags behind the sota, which exploits in-domain category-specific training, in particular the excellent fine-tuned person detectors, which lead to superior detection accuracy (DetA in Table~\ref{tab:mot17}).
This is evident from a 24\% DetA drop of OC-SORT using a public human detector.
Meanwhile, comparable association accuracy to that of OC-SORT$^*$ indicates the effectiveness of the consolidation module responsible for association.

\begin{table*}
\centering

\begin{minipage}[t]{0.45\textwidth}
\centering
\caption{Tracking performance on the MOT17. The colors denote UGO performing \textcolor{softgreen}{better}/\textcolor{softorange}{worse} compared to the individual method, $^*$ denotes public detection setup.}
\resizebox{\linewidth}{!}{
\begin{tabular}{lccc}
\toprule
Method & HOTA  & DetA  & AssA \\
\hline
MPNTrack17 & 46.6 \textcolor{softgreen}{(7\%)} & 46.2 \textcolor{orange}{(-2\%)} & 47.3 \textcolor{softgreen}{(18\%)} \\
eTC17 & 45.1 \textcolor{softgreen}{(11\%)} & 44.1 \textcolor{softgreen}{(3\%)} & 46.4 \textcolor{softgreen}{(19\%)} \\
Tracktor++v2 & 45.1 \textcolor{softgreen}{(11\%)} & 45.3 \textcolor{gray}{(0\%)} & 45.0 \textcolor{softgreen}{(24\%)} \\
\hline
ByteTrack~\cite{zhang2022bytetrack} & 63.1 \textcolor{orange}{(-21\%)} & \textbf{64.5} \textcolor{orange}{(-30\%)} & \textbf{62.0} \textcolor{orange}{(-10\%)} \\
OC-Sort~\cite{OC-SORT} & \textbf{63.2} \textcolor{orange}{(-21\%)} & 63.2 \textcolor{orange}{(-28\%)} & 63.4 \textcolor{orange}{(-12\%)} \\
OC-Sort$^*$~\cite{OC-SORT} & 54.6 \textcolor{orange}{(-8\%)} & 47.8 \textcolor{orange}{(-5\%)} & 57.6 \textcolor{orange}{(-3\%)} \\
\hline
\name{} & 50.0 & 45.3 & 55.6  \\
\bottomrule
\end{tabular}}
\label{tab:mot17}
\end{minipage}
\hfill
\begin{minipage}[t]{0.5\textwidth}
\centering
\caption{Video object counting results on Science-Count~\cite{countvid}. Methods marked with * accept text prompts instead of exemplars.}
\resizebox{\linewidth}{!}{
\begin{tabular}{lcccc}
\toprule
 & \multicolumn{2}{c}{Penguins} & \multicolumn{2}{c}{Crystals} \\
\cline{2-5}
Method & MAE↓ & RMSE↓ & MAE↓ & RMSE↓ \\
\hline
GDINO$_\text{MASA}$*~\cite{liu2024grounding} & 9.0 & 11.5 & 72.3 & 87.9 \\
CountGDB$_\text{ByteTrack}$*~\cite{countvid} & 4.3 & 5.5 & 71.6 & 88.1 \\
CountVid~\cite{countvid}* & 4.0 & 5.3 & 69.1 & 86.0 \\
\hline
CountGDB$_\text{ByteTrack}$~\cite{countvid} & 4.0 & 4.2 & 31.1 & 52.8 \\
GeCo$_\text{SAM2.1}$~\cite{geco} & 11.3 & 14.5 & 46.1 & 82.8 \\
CountVid~\cite{countvid} & 3.3 & 4.8 & 33.7 & 59.8 \\
\name{}& \textbf{2.7} & \textbf{3.5} & \textbf{24.6} & \textbf{43.4} \\
\bottomrule
\end{tabular}}
\label{tab:sciencecount}
\end{minipage}

\end{table*}

\subsection{Comparison with video object counters}
We finally evaluate \name{} in video object counting performance on the recent ScienceCount benchmark~\cite{countvid}. 
The MAE and RMSE~\cite{countvid}, computed by counting trajectories without final confirmation-based filtering, are reported in Table~\ref{tab:sciencecount}. 
\name{} outperforms the sota 
CountVid~\cite{countvid} by 18\%/27\% in MAE and 25\%/27\% in RMSE on the Penguins and Crystals subsets, respectively.
Notably, CountVid is a video counting specialist, employing a multistage, forward-backward tracking pipeline with separate detection and tracking backbones. While \name{} processes the video in a single pass, its dominance over the counting specialist emphasizes the strong generalization capabilities.

\subsection{Ablation study}

\name{} design choices are analyzed on GMOT-40~\cite{bai2021gmot} in Table~\ref{tab:ablation} and further in supplementary material.

\begin{table*}
\centering
  \caption{\name{} ablation on GMOT-40~\cite{bai2021gmot}. }
  \label{tab:ablation}
\resizebox{\textwidth}{!}{
\setlength{\tabcolsep}{6pt} 
\begin{tabular}{lcccccccccccc}
\toprule
Method & HOTA & DetA & AssA & MOTA & FN & FP & IDSw & MT & PT & ML & FM & IDF1 \\
\midrule
\name{}$_{\overline{\text{CONS}}}$ &  50.98&49.11 & 54.36 &43.12 &  61835 &59794 &24173 &1084 & 671& 189& 8414 & 61.47   \\
\name{}\textsubscript{HM} & 52.35 &39.42 &70.55  & 5.88 & 34457 & 204395 &2402 &1460 & 328& 156 & 4553 & 60.76   \\
\name{}$_{\overline{\Theta_1}}$ & 52.39 & 55.56& 50.48 &57.11& 39900 &64272  & 5773 & 1360 &421 & 163 & 4478 & 56.86 \\
\name{}\textsubscript{D4S} &61.00 & 53.80& 70.49 &57.33&51892 & 55349 & 2132 &1252 &430 & 262&  4409 & 72.81    \\
\name{}$_{\overline{\Theta_2}}$ &62.76&55.55 & 72.11  & 59.16 & 40059 & 63251 & 1375 & 1351& 419   &    174      & 4633 & 75.66 \\
\name{}$_{\overline{\text{DRM}}}$  &63.19  & 56.23 & 72.20 &60.30  & 41057 & 59463 & 1244 &1331 & 434 &179  & 4554 &76.10  \\
\name{}$_{\text{CAL}}$ &63.73 & 56.57 & 72.98 & 61.03 & 40126 & 58646 & 1121 & 1344 & 423 & 177 & 4501 & 76.89 \\

\name$_{\text{1ex}}$  &62.37 & 54.53 & 72.54 & 58.34 & 48908 & 56773 & 1113 & 1305 & 417 & 222 & 4406 & 75.07 \\
\name$_{\text{4-rand-GT}}$  &63.17 & 55.52 & 73.13 & 58.84 & 37425 & 66933 & 1139 & 1372 & 429 & 143 & 4624 & 76.34 \\
\midrule
\name{}  &63.83 & 56.72 & 73.03 & 61.39 & 39670 & 58165 & 1136 & 1349 & 425 & 170 & 4473 & 77.09 \\
\bottomrule
\end{tabular}}
\end{table*}

{\bf Consolidation module}. 
Several variations are considered (Table~\ref{tab:ablation}). 
The first, \name{}\textsubscript{$\overline{\mathrm{CONS}}$}, has the consolidation module removed, and each detection is associated by the instance tracker with highest corresponding IoU. 
\name{}\textsubscript{$\overline{\mathrm{CONS}}$} results in 
 20\% HOTA and 30\% MOTA drops compared to \name{} and nearly $2 \times$ as many trajectory fragmentations,  
due to missed detections and weak temporal consistency.
Next, the consolidation module is replaced by Hungarian matching~\cite{hun_matching} (\name{}\textsubscript{HM}) for optimal assignment between tracker and detector masks.   
\name{}\textsubscript{HM} results in 18\% HOTA and 90\% MOTA performance drops. The results verify the importance and robustness of the proposed consolidation module, which standard MOT-style matching strategies cannot replace. 

{\bf Calibration of detector and tracker logits.} 
Consolidation relies on direct competition between tracker and detector logits, both computed with the shared SAM2 mask decoder; we verify their compatibility by scaling detector logits to match the tracker’s mean amplitude (\name{}$_{\text{CAL}}$), on GMOT-40. 
Performance remains unchanged (Table~\ref{tab:ablation}), indicating that no additional calibration is required.


{\bf Consolidation module potentials}.
The potentials $\Theta_{1}(\cdot)$ and $\Theta_{2}(\cdot)$ in (\ref{eq:loss1}) control the consolidation labeling dynamics.
Removing $\Theta_{1}(\cdot)$, denoted by \name{}\textsubscript{$\overline{\Theta_{1}}$} decreases HOTA by 18\% and MOTA by 7\%, while removing $\Theta_{2}(\cdot)$ leads to 2\% HOTA and 4\% MOTA reduction (denoted by \name{}\textsubscript{$\overline{\Theta_{2}}$}). 
$\Theta_{1}(\cdot)$ is thus central for resolving competition, favoring trackers over detections and reinforcing detector-supported tracks, 
while $\Theta_{2}(\cdot)$ suppresses uncertain or unstable masks.
Together, they ensure stable, pixel-level consistent identities even in cluttered and highly dynamic scenes.
This validates the importance of the proposed potentials for strong performance.

{\bf Instance-level memory}. Replacing the proposed instance memory (Section~\ref{sec:method_memory}) by the original SAM2~\cite{sam2} 
(\name{}\textsubscript{$\overline{\text{DRM}}$}) results in 2\% MOTA and 1\% HOTA drops, 10\% more identity switches and more fragmentations.
This supports importance of the proposed memory for identity stability.
Replacing our memory management with that of~\cite{dam4sam} (\name{}\textsubscript{D4S}) leads to  4\% HOTA and 7\% MOTA drops, confirming that the proposed memory management offers a more robust multi-object tracking. 


{\bf Category-level memory}. Removing FIFO and keeping only the initial exemplar in the memory (\name{}\textsubscript{1ex}), leads to 5\% MOTA and 2\% HOTA drop (Table~\ref{tab:ablation}), indicating that adaptive exemplar updates not only enhance instance discovery but also improve long-term identity consistency by stable trajectories mining. Constructing memory from four randomly selected ground-truth boxes from the first frame and keeping it fixed throughout the sequence (\name{}\textsubscript{4-rand-GT}) leads to 4\% MOTA and 1\% HOTA drops, further emphasizing the benefits of our on-the-fly trajectory mining.

\section{Conclusion}
We introduced \name{}, a unified general multi-object tracker that integrates exemplar-conditioned detection and mask-based instance propagation within a single architecture. 
A novel consolidation module resolves conflicting and over-segmented outputs, yielding mutually exclusive masks, while hierarchical memory enables robust global and per-instance modeling.
\name{} achieves state-of-the-art performance in GMOT and video counting, while remaining competitive with specialist MOT methods, thus confirming the potential of the new GMOT design paradigm.

\section{Acknowledgements}
This work was supported by the Slovenian Research Agency program P2-0214 and project J2-60054, as well as the supercomputing network SLING (ARNES, EuroHPC Vega - IZUM), and the Slovenian Ministry of MESY and EC/EuroHPC JU via the project SLAIF (grant number 101254461).





{
    \small
    \bibliography{main}
}


\appendix

\section{The mask consolidation algorithm}
\label{sec:method_consolidation_derivation}

We provide further details on derivation of the UGO mask consolidation algorithm for the detector and tracker masks. 
We first introduce the notations, then define the original cost function and the surrogate loss, and then proceed to derive the corresponding minimization algorithm.

Let $\Omega=\{1,\dots,H\}\times\{1,\dots,W\}$ be the pixel domain and let $S=N_{\mathcal D}+N_{\mathcal T}+1$ denote the number of labels, i.e., potentially competing masks for explaining individual pixels (detections, trackers, and background).
We form a logit tensor
\begin{equation}
\mathbf L \in \mathbb R^{H\times W\times S}, \qquad 
\end{equation}
by concatenating all detector logits, all tracker logits, and a constant background logit map (i.e., $\lambda_\mathrm{BG}$). Let $\mathbf{L}(\mathbf{x},s)$ denote the value of the tensor at pixel location $\mathbf x$ for label $s$.

We seek a mutually-exclusive pixel labeling $\mathbf Y:\Omega\to\{1,\dots,S\}$,
with $\mathbf Y(\mathbf x)$ reading out the label assigned to a pixel $\mathbf x$.
Further, let $N_s(\mathbf Y)$ be a function that counts the number of pixels assigned a label $s$ in the labeling $\mathbf Y$,
\begin{equation}
N_s(\mathbf Y) =  \sum_{\mathbf x\in\Omega}\mathbf 1_{[\mathbf Y(\mathbf x)\equiv s]},
\end{equation}
where $\mathbf Y(\mathbf x)\equiv s$ verifies that the label at pixel $\mathbf x$ is equal to $s$ in the labeling $\mathbf Y$.


\subsection{The labeling cost function}

To enforce a desired behavior of the optimization (as discussed in the paper, Section~3.3), we introduce two label-wise potentials $\Theta_1(s;\mathbf Y)$ and $\Theta_2(s;\mathbf Y)$ -- for brevity we will omit $\textbf Y$ in the following, i.e.,  $\Theta_1(s)$ and $\Theta_2(s)$. 
We define these potentials to encode:
(i) a preference for trackers over detections, (ii) a preference for trackers with high detector support, and 
(iii) suppression of unstable labels whose final mask deviates from its initial mask. 
Concretely, we make the potentials depend on $\mathbf Y$ through IoU terms between masks induced by $\mathbf Y$ and (binarized) initial masks (see Section~3.3 in the paper for definitions). 
The labeling cost function to be optimized is thus
\begin{equation} 
    \label{eq:loss1_}
        \hat{\mathcal{L}}(\mathbf{Y}) = 
        -\sum_{\mathbf{x}\in \Omega \mathrm{ ; } s = \mathbf{Y}(\mathbf{x})} \log\Big( \mathbf{L}(\mathbf{x},s) \cdot \Theta_1(s) \cdot \Theta_2(s) \Big),
\end{equation}
 {where $\mathbf{L}(\mathbf{x},s)>0$ always holds, as the label set includes a background label with a constant nonnegative logit $\lambda_{\mathrm{BG}}>0$. Consequently, any detector or tracker label on a non-positive logit cannot be selected over the background. In practice, the IoU-based potentials are also lower-bounded by a small $\epsilon>0$, ensuring $\Theta_1(s)\Theta_2(s)>0$ and making the logarithm well-defined.}

\subsection{Reweighted surrogate objective}

Due to nonlinear interaction between pixel labeling in $\Theta_1(s)$ and $\Theta_2(s)$, the objective (\ref{eq:loss1_}) does not adhere to simple optimization over $\mathbf Y$. Thus, auxiliary variables $A_1(s)$ and $A_2(s)$ are introduced for $\Theta_1(s, \mathbf Y)$ and $\Theta_2(s, \mathbf Y)$ that decouple the logits from the latter and make pixel labeling fully separable. This leads to the following surrogate objective

\begin{eqnarray}
\label{eqn:loss_aux}
{\mathcal{L}}(\mathbf{Y}, A_1, A_2) = \nonumber \\ 
-\sum_{\mathbf x\in\Omega}\Big(
\log (\mathbf L(\mathbf x,\mathbf Y(\mathbf x)) A_1(\mathbf Y(\mathbf x)) A_2(\mathbf Y(\mathbf x)))
\Big) 
\nonumber \\
+\sum_{s=1:S} N_s 
\Big(
\big(\log A_1(s)-\log \Theta_1(s)\big)^2 + \nonumber \\
\big(\log A_2(s)-\log \Theta_2(s)\big)^2
\Big),
\label{eq:Surrogate}
\end{eqnarray}
where the $N_s$ is introduced to remove the influence of the mask size, i.e., such that masks corresponding to large objects are not preferred over those of small objects.

This objective can be minimized by an iterated scheme that exchanges between estimation of $A_1(s)$ and $A_2(s)$ and estimation of $\mathbf Y$, closely following the standard Majorize-Minorize~\cite{mm} approach. 
In particular, the following steps are exchanged:\\
\noindent \textbf{Step 1} assumes an estimate $\mathbf{Y}^{(k-1)}$ from previous iteration and solves the following minimization:
\begin{equation}
     A_1^{(k)}, A_2^{(k)} = \arg \min_{ A_1, A_2 }  \mathcal{L}(\mathbf{Y}^{(k-1)}, A_1, A_2).
\end{equation}
\\ \noindent \textbf{Step 2} then fixes $A_1^{(k)}$ and $A_2^{(k)}$, and solves the following minimization:
\begin{equation}
     \mathbf{Y}^{(k)} = \arg \min_{ \mathbf{Y} }  \mathcal{L}(\mathbf{Y}, A_1^{(k)}, A_2^{(k)}).
\end{equation}

\subsection{Step 1 minimization problem}

The objective (\ref{eq:Surrogate}) is rewritten into 
\begin{align}
{\mathcal{L}}(\mathbf{Y}^{(k-1)}, A_1, A_2) = \sum_{ s = 1 : S }\Big( \\  {-}
    \sum_{\mathbf x \in \Omega}\Big[
        \log(A_1(s)A_2(s)) 1_{[\mathbf Y^{(k-1)}(\mathbf x)\equiv s]} + \varepsilon(s)
        \Big]
    &\\
    + {N}_s^{(k-1)}\Big( (\log A_1(s)-\log  \Theta_1^{(k-1)}(s))^2
    &\\
    + (\log A_2(s)-\log \Theta_2^{(k-1)}(s))^2\Big)
    \Big),
\label{eq:Surrogate2}
\end{align}
where $\varepsilon(s)$ absorbs the terms not depending on $A_1$ and $A_2$, while $\Theta_1^{(k-1)}$, $\Theta_1^{(k-1)}$ and ${N}_s^{(k-1)}$ are evaluated at $\mathbf Y^{(k-1)}$.

Then the updates for $A_1^{(k)}(s)$ and $A_2^{(k)}(s)$ are obtained by differentiating the objective and setting to zero, i.e.,
\begin{align}
    \frac{ \partial {\mathcal{L}}(\mathbf{Y}^{(k-1)}, A_1(s), A_2(s))   }{\partial  {\log}A_1(s)} \equiv 0 \label{eq:deriv_a1} \\
    \frac{ \partial {\mathcal{L}}(\mathbf{Y}^{(k-1)}, A_1(s), A_2(s))   }{\partial  {\log}A_2(s)} \equiv 0 . \label{eq:deriv_a2}
\end{align}
The derivative in~(\ref{eq:deriv_a1}) is defined as
\begin{equation}
    \frac{ \partial {\mathcal{L}}   }{\partial  {\log}A_1(s)} = -N_s + 2  {N_s}\log A_1(s) -2N_s\log \Theta_1^{(k-1)}(s),
\end{equation}

which gives the following update for $A_1^{(k)}(s)$
\begin{equation}
      A_1^{(k)}(s) = \Theta^{(k-1)}_1(s) \cdot e^{1/2},
\end{equation}
while a similar derivation from~(\ref{eq:deriv_a2}) with  $\frac{ \partial {\mathcal{L}}   }{\partial  {\log}A_2(s)} \equiv 0$ gives
\begin{equation}
      A_2^{(k)}(s) = \Theta^{(k-1)}_2(s) \cdot e^{1/2}.
\end{equation}

\subsection{Step 2 minimization problem}

In the second step, the auxiliary variables are fixed, the labeling $\mathbf Y$
can be obtained from (\ref{eqn:loss_aux}) by minimizing 
\begin{align}
     \mathbf{Y}^{(k)} &= \arg \min_{ \mathbf{Y} }  \mathcal{L}(\mathbf{Y}, A_1^{(k)}, A_2^{(k)})\\
      &= \arg \max_{ \mathbf{Y} } \sum_{\mathbf x\in\Omega ; s=\mathbf Y(x)} L(\mathbf x, s) A_1^{(k)}(s) A_2^{(k)}(s),\nonumber
\end{align}
which decomposes over pixels, thus the labels can be computed for each pixel $\mathbf x$ separately:  
\begin{align}
     \mathbf{Y}^{(k)}(\mathbf{x}) = \arg \max_{ \mathbf{Y}(\mathbf{x}) }  L(\mathbf x, \mathbf{Y}(\mathbf{x})) A_1^{(k)}(\mathbf{Y}(\mathbf{x})) A_2^{(k)}(\mathbf{Y}(\mathbf{x})).
\end{align}

\section{HAM Architecture Details}

\begin{figure}[h!]
  \centering
   \includegraphics[width=\linewidth]{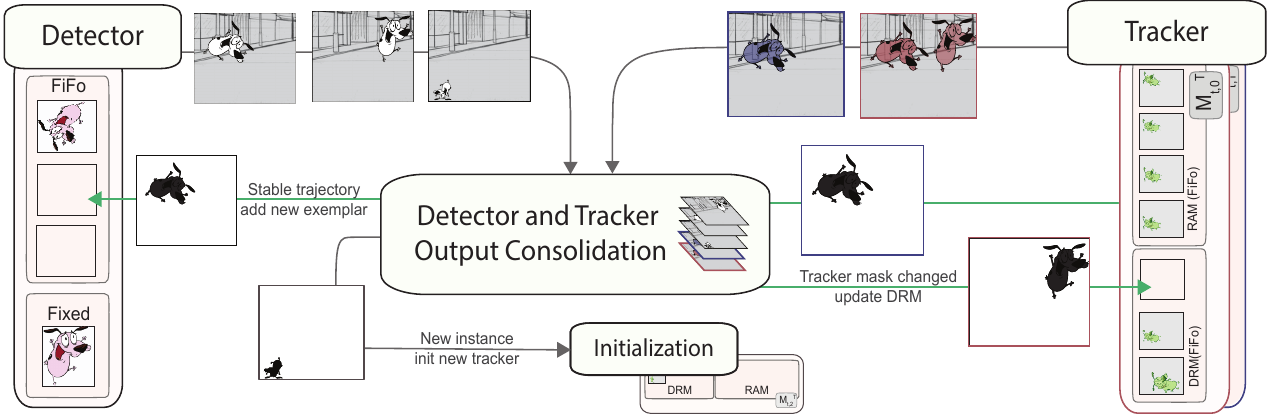}
   \caption{HAM overview: stable tracks provide robust updates for detector category-level memory, while instance-level memories are updated robustly by the result of consolidation analysis.}
   \label{fig:mem_update}
\end{figure}

The hierarchical adaptive memory (HAM), described in Section~\ref{sec:method_memory}, consists of two complementary components: a category-level memory $\mathcal{M}^D_t$ used for exemplar-conditioned detection, and per-instance memories $\{\mathcal{M}^T_{t,k}\}$ used for instance propagation, see Figure~\ref{fig:mem_update}.
The category-level memory maintains a compact set of reliable exemplars, while each instance-level memory is decomposed into distractor-resolving (DRM) and recent appearance (RAM) buffers, enabling robust discrimination and temporal adaptation. 
The two levels are tightly coupled through consolidation: reliable tracks update $\mathcal{M}^D_t$, while consolidation-driven corrections trigger updates of $\mathcal{M}^T_{t,k}$, ensuring consistent global and instance-level representations.

\section{Experimental Evaluation on Animal Track}

\begin{table*}
\centering
\caption{State-of-the-art comparison on AnimalTrack~\cite{animaltrack}.}
\resizebox{\textwidth}{!}{
\setlength{\tabcolsep}{3pt}
\begin{tabular}{lccccccccccccccc}
\toprule
Method & HOTA$\uparrow$ & MOTA$\uparrow$ & IDF1$\uparrow$ & IDP$\uparrow$ & IDR$\uparrow$ & MT$\uparrow$ & PT & ML$\downarrow$ & FP$\downarrow$ & FN$\downarrow$ & Pr$\uparrow$ & Re$\uparrow$ & F1$\uparrow$ & IDSw$\downarrow$ & FM$\downarrow$ \\
\midrule

JDE & 26.8 & 27.3 & 31.0 & 51.0 & 22.0 & 106 & 414 & 584 & 17887 & 155623 & 0.830 & 0.360 & 0.502 & 3187 & 5031 \\

FairMOT & 30.6 & 29.0 & 38.8 & 62.8 & 28.0 & 143 & 462 & 499 & 17653 & 152624 & 0.837 & 0.372 & 0.515 & 2335 & 5447 \\

Trackformer & 31.0 & 20.4 & 36.5 & 40.9 & 32.8 & 230 & 491 & 383 & 70404 & 118724 & 0.639 & 0.512 & 0.568 & 4355 & 3725 \\

TADAM & 32.5 & 36.5 & 37.2 & 44.4 & 32.0 & 258 & 495 & 351 & 41728 & 110048 & 0.761 & 0.547 & 0.637 & 2538 & 4469 \\

DeepSORT & 32.8 & 41.4 & 35.2 & 49.7 & 27.2 & 213 & 452 & 439 & \textbf{14131} & 124747 & 0.893 & 0.487 & 0.630 & 3503 & 4527 \\

ByteTrack & 40.1 & 38.5 & 51.2 & 64.9 & 42.3 & 310 & 465 & 329 & 31591 & 116587 & 0.800 & 0.521 & 0.631 & 1309 & 3513 \\

IOUTrack & 41.6 & 55.7 & 45.7 & 51.9 & 40.7 & 388 & 454 & 262 & 25206 & 77847 & 0.868 & 0.680 & 0.762 & 4639 & 5259 \\

SORT & 42.8 & 55.6 & 49.2 & 58.5 & 42.4 & 333 & 470 & 301 & 19099 & 86257 & 0.891 & 0.645 & 0.749 & 2530 & 3730 \\

OMC & 43.0 & 53.4 & 50.3 & 61.8 & 42.4 & 324 & 478 & 302 & 15910 & 92570 & \textbf{0.904} & 0.619 & 0.735 & 4938 & 7162 \\

Tracktor++ & 44.2 & 55.2 & 51.0 & 58.5 & 45.1 & 364 & 472 & 268 & 25477 & 81538 & 0.864 & 0.665 & 0.751 & 1976 & 4149 \\

TransTrack & 45.4 & 48.3 & 53.4 & 63.4 & 46.1 & 327 & 416 & 361 & 28553 & 95212 & 0.838 & 0.608 & 0.705 & 1978 & 6459 \\

QDTrack & 47.0 & \textbf{55.7} & 56.3 & 65.6 & 49.3 & 367 & 420 & 317 & 22696 & 83057 & 0.876 & 0.658 & 0.752 & 1970 & 5656 \\

SAM3 & 52.1 & 43.9 & 66.4 & 68.4 & 64.4 & 642 & 280 & 188 & 60979 & 74880 & 0.734 & 0.692 & 0.713 & 470 & 2588 \\

\name{} & \textbf{59.4} & 53.8 & \textbf{71.7} & \textbf{67.6} & \textbf{76.3} & \textbf{700} & 254 & \textbf{156} & 71695 & \textbf{40157} & 0.739 & \textbf{0.835} & \textbf{0.783} & \textbf{463} & \textbf{2490} \\

\bottomrule
\end{tabular}}
\label{tab:animal_full_prf}
\end{table*}

Table~\ref{tab:animal_full_prf} provides extended results on AnimalTrack, including false positives (FP), false negatives (FN), precision, recall, and F1 score (Table~\ref{tab:animal_full_prf}).
The results show that \name{} besides achieving the highest HOTA, it achieves the highest recall and F1 score among all compared trackers. 
In particular, \name{} reduces FN (40157 FN) at least two-fold compared to all specialist methods, including QDTrack (83057 FN), demonstrating significantly better temporal coverage of target instances. 
Although \name{} produces more false positives than some specialist detectors (e.g., OMC or QDTrack), it achieves the strongest overall F1 performance. 
This confirms that \name{} favors consistent instance coverage and identity preservation over overly conservative detection suppression.

\paragraph{Comparison with SAM3 (text-prompted MOT).}
We additionally evaluate SAM3~\cite{sam3} on AnimalTrack. 
Unlike \name{}, SAM3 does not support exemplar-conditioned multi-object tracking in the GMOT sense, where a single first-frame exemplar defines the category for the entire video. While SAM3 allows image exemplars as prompts, these are applied within the same frame (e.g., for the detection of all objects of the same category in that frame) and are not designed, nor do they work, as a persistent cross-frame category specification.
We therefore evaluate SAM3 using a text prompt specifying the ground-truth category for each sequence, which aligns with its standard usage in video tracking.

Text prompting offers a semantic advantage over exemplar-based specification, particularly for fine-grained categories. 
A textual label such as ``duck'' or ``goose'' explicitly defines the target class, whereas a few visual exemplars may not clearly indicate whether only that species or also visually similar birds should be tracked. 
For example, bird species can appear highly similar in crowded scenes, making the category boundary ambiguous from a few exemplars. 
Moreover, SAM3 is trained on large-scale image and video segmentation data spanning a broad set of visual concepts, which supports strong generalization under semantic (text) prompting. In contrast, the few-shot detector used in \name{} is not explicitly trained for the target categories and must rely solely on limited exemplar supervision. Despite this advantage, \name{} outperforms SAM3 on the primary tracking metrics.
This exposes an important constraint/limitation of \name{}. 
Since the target category is specified only through visual exemplars, \name{} cannot explicitly control the semantic granularity of the category to be tracked, which is inherited from GECO2 detector~\cite{geco2}.

\begin{figure}
  \centering
   \includegraphics[width=\linewidth]{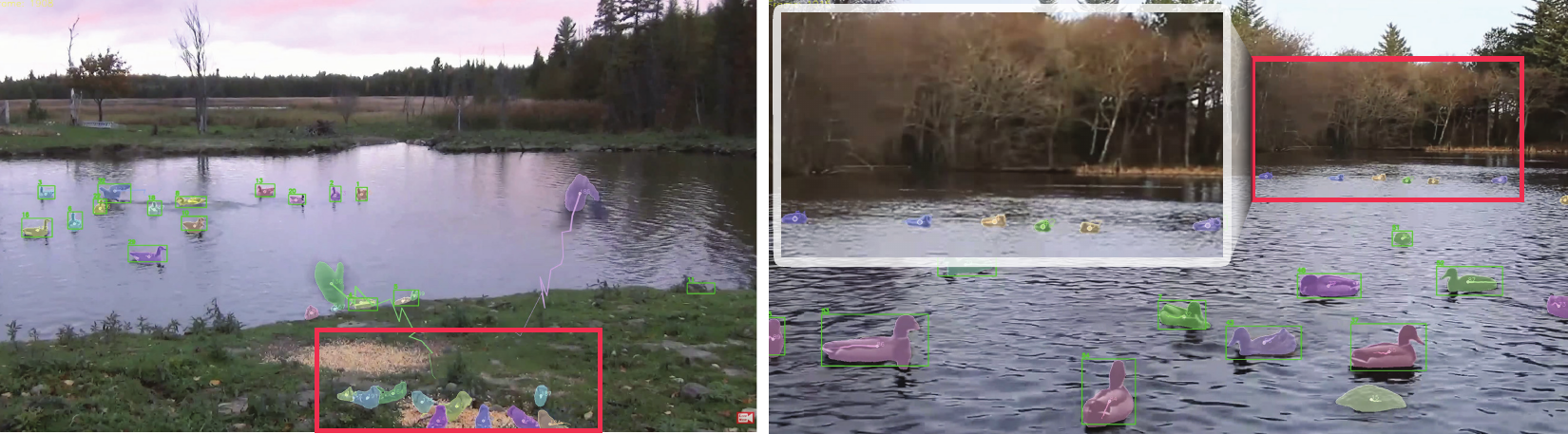}
   \caption{
Two sequences from AnimalTrack where UGO produces the most false detections. Red bounding boxes are overlaid to highlight the relevant regions. Ground-truth bounding boxes are shown in green -- note that some instances are missing in the ground truth annotations.
   }
   \label{fig:animal_track}
\end{figure}

In contrast, \name{} must infer the category solely from visual evidence contained in a few exemplars, without access to explicit semantic supervision, which is a hard task in fine-grained classification Figure~\ref{fig:animal_track}.
Despite this advantage, \name{} outperforms SAM3 on the primary tracking metrics. 
\name{} outperforms SAM3 in HOTA by 14\% and a remarkable 23\% MOTA.
While SAM3 produces fewer identity switches and fragmentations, it is because of lower detection recall (20\% lower relatively) and reduced overall instance coverage. 
These results highlight that the proposed exemplar-driven consolidation and memory mechanism provides stronger identity consistency and more complete tracking, even without access to explicit semantic category labels. 

For evaluating SAM3~\cite{sam3}, we use the default parameters of the official video predictor. Table~\ref{tab:sam3_video_thresholds} summarizes these thresholds. SAM3 uses a loose association threshold to avoid duplicate masklet creation, while a stricter threshold determines whether an existing masklet is sufficiently supported by detector evidence. Newly initialized masklets are further filtered during a hotstart period, in which unmatched masklets or those repeatedly duplicating existing trajectories are removed.
In contrast, UGO omits many of these thresholds by using the proposed consolidation module, since duplicate handling, new instance discovery, even mask correction, and instance memory update logic are handled in a single unified step.

\begin{longtable}{p{0.38\linewidth} p{0.13\linewidth} p{0.41\linewidth}}
\caption{Threshold summary for SAM3~\cite{sam3}.}
\label{tab:sam3_video_thresholds}
\\
\toprule
\textbf{Parameter} & \textbf{Default} & \textbf{Role} \\
\midrule
\endfirsthead

\toprule
\textbf{Parameter} & \textbf{Default} & \textbf{Role} \\
\midrule
\endhead

\bottomrule
\endlastfoot



\texttt{assoc\_iou\_thresh}
& $0.1$
& Loose detection-to-track IoU threshold for deciding whether a detection is new.
\\

\texttt{trk\_assoc\_iou\_thresh}
& $0.5$
& Stricter IoU threshold for deciding whether a tracker masklet is matched.
\\

\texttt{new\_det\_thresh}
& $0.7$
& Minimum detector score required to spawn a new object.
\\

\texttt{HIGH\_CONF\_THRESH}
& $0.8$
& Threshold for high-confidence detector reconditioning.
\\

\texttt{iou\_thresh\_recondition}
& $0.8$
& Default IoU threshold for detector-to-tracker memory refresh.
\\
\texttt{suppress\_overlapping\_based}\\\texttt{\_on\_recent\_occlusion\_threshold}
& $0.7$
& IoU threshold for suppressing highly overlapping tracker masks based on recent occlusion history.
\\
\texttt{detector\_tracker\_confirmation}
& $0.5$
& Ratio of detector-confirmed tracker masks required to keep a masklet; otherwise, the masklet is removed.
\\
\texttt{hotstart\_delay}
& $15$
& Number of frames used to judge early masklet stability.
\\

\texttt{hotstart\_unmatch\_thresh}
& $8$
& Number of unmatched frames after which a new masklet is removed.
\\

\texttt{hotstart\_dup\_thresh}
& $8$
& Number of duplicated frames before hotstart removal.
\\

\texttt{max\_trk\_keep\_alive}
& $30$
& Maximum keep-alive counter.
\\

\texttt{recondition\_every\_nth\_frame}
& $16$
& Periodic detector-based tracker refresh interval.
\\

\texttt{masklet\_confirmation}\\\texttt{\_consecutive\_det\_thresh}
& $3$
& Number of detections required if confirmation is enabled.
\\

\texttt{mask logit threshold}
& $0$
& Detector/tracker mask logits are binarized with threshold $>0$.
\\

\texttt{NO\_OBJ\_LOGIT}
& $-10$
& Logit assigned to suppressed masks before memory encoding.
\\

\end{longtable}

\section{Ablation Study}  \label{sec:abl-sensitivity}

This section analyzes the sensitivity of \name{} to the
\emph{conservative gates} used by the consolidation module and the memory/lifecycle rules.
These gates control when masks are accepted, suppressed, or used for memory updates.
All experiments are conducted on GMOT-40~\cite{bai2021gmot} by varying one gate at a time
while keeping the remaining gates fixed to the default configuration.
Results are reported in Table~\ref{tab:ablation_suppl}.

\begin{table*}[ht!]
\centering
  \caption{\name{} ablation on GMOT-40~\cite{bai2021gmot}. }
  \label{tab:ablation_suppl}
\resizebox{\textwidth}{!}{
\setlength{\tabcolsep}{3pt} 
\begin{tabular}{lcccccccccccc}
\toprule
Method & HOTA & DetA & AssA & MOTA & FN & FP & IDSw & MT & PT & ML & FM & IDF1 \\
\midrule
$\theta_0=0.3$ &63.71 & 56.60 & 72.91 & 61.12 & 40195 & 58347 & 1130 & 1348 & 426 & 170 & 4494 & 76.85 \\
$\theta_0=0.4$ &63.75 & 56.62 & 72.96 & 61.15 & 40000 & 58416 & 1165 & 1345 & 431 & 168 & 4500 & 76.92 \\
$\theta_0=0.6$ &63.82 & 56.65 & 73.09 & 61.20 & 39769 & 58605 & 1092 & 1353 & 422 & 169 & 4463 & 77.03 \\
$\theta_0=0.7$ &63.77 & 56.64 & 72.98 & 61.20 & 39695 & 58666 & 1107 & 1344 & 433 & 167 & 4487 & 77.00 \\
$\theta_0=0.8$ &63.62 & 56.57 & 72.72 & 60.98 & 39996 & 58942 & 1082 & 1347 & 431 & 166 & 4492 & 76.78 \\
\midrule
$\tau_{\text{md}}=0.45$  &63.45 & 56.33 & 72.70 & 60.37 & 38370 & 61941 & 1269 & 1359 & 419 & 166 & 4575 & 76.49 \\
$\tau_{\text{md}}=0.5$  &63.66 & 56.45 & 73.01 & 60.74 & 38617 & 60793 & 1219 & 1357 & 422 & 165 & 4532 & 76.80 \\
$\tau_{\text{md}}=0.55$   &63.67 & 56.51 & 72.96 & 60.76 & 39616 & 59756 & 1212 & 1354 & 420 & 170 & 4506 & 76.75 \\
 $\tau_{\text{md}}=0.65$  &63.73 & 56.66 & 72.86 & 61.32 & 40446 & 57620 & 1091 & 1342 & 426 & 176 & 4492 & 76.90 \\
$\tau_{\text{md}}=0.7$  &63.56 & 56.57 & 72.59 & 61.45 & 41299 & 56408 & 1103 & 1331 & 434 & 179 & 4466 & 76.73 \\
\midrule
$\tau_{\text{lo}}=0.1$  &63.78 & 56.67 & 72.99 & 61.14 & 39461 & 58990 & 1171 & 1351 & 427 & 166 & 4469 & 76.98 \\
$\tau_{\text{lo}}=0.3$ &63.71 & 56.72 & 72.76 & 61.36 & 39771 & 58148 & 1137 & 1350 & 427 & 167 & 4469 & 76.88 \\
\midrule
 $\tau_{\text{hi}}=0.75$&63.65 & 56.65 & 72.69 & 61.29 & 40050 & 58023 & 1162 & 1341 & 437 & 166 & 4488 & 76.81 \\
$\tau_{\text{hi}}=0.85$  &63.72 & 56.66 & 72.87 & 61.19 & 39762 & 58613 & 1118 & 1346 & 426 & 172 & 4482 & 76.82 \\
$\tau_{\text{hi}}=0.95$  &63.26 & 56.25 & 72.32 & 60.43 & 40144 & 60185 & 1108 & 1356 & 405 & 183 & 4340 & 76.19 \\
\midrule

\name{}  & 63.83 & 56.72 & 73.03 & 61.39 & 39670 & 58165 & 1136 & 1349 & 425 & 170 & 4473 & 77.09 \\
\bottomrule
\end{tabular}}
\end{table*}

\paragraph{Tracker preference strength $\theta_0$.}
The constant $\theta_0$ controls the preference of trackers over detector logits in $\Theta_1(\cdot)$, and it also sets the scale of the detector-support term that boosts tracker masks consistent with detections.
We evaluate $\theta_0\in\{0.3, 0.4, 0.6, 0.7\}$ and compare it with the default setting 0.5 in \name{}.
As shown in Table~\ref{tab:ablation_suppl}, performance is essentially unchanged across this range: HOTA varies by less than $0.21$ points around the default, and MOTA remains within $\approx 0.41$ points of the default (with the largest drop observed at $\theta_0=0.8$ due to increased FP). This indicates that the consolidation labeling is insensitive to the exact strength of preference, provided that trackers are moderately favored.

\paragraph{Overlap threshold $\tau_{\mathrm{md}}$.}
The moderate threshold $\tau_{\mathrm{md}}$ is used for:
(i) deciding whether a tracker mask is \emph{confirmed} by any detector that gates trajectory verification, and (ii) initialization of new tracks.
We evaluate $\tau_{\mathrm{md}}\in\{0.45,0.5,0.55,0.65,0.7\}$.
Lowering $\tau_{\mathrm{md}}$ relaxes confirmation, which increases false positives; this is reflected by the drop of $2\%$ MOTA at $\tau_{\mathrm{md}}=0.45$.
Increasing $\tau_{\mathrm{md}}$ makes confirmation stricter; at $\tau_{\mathrm{md}}=0.7$ it starts to reduce MT and increase ML, rejecting correct trajectories as invalid. Overall, $\tau_{\mathrm{md}}=0.6$ is the stable balance, yielding the least ML.

\paragraph{Tracker-collapse threshold $\tau_{\mathrm{lo}}$.}
After consolidation, labels whose final mask is inconsistent with the initial mask are removed.  
For trackers, a permissive threshold $\tau_{\mathrm{lo}}$ is used to avoid terminating tracks whose masks were corrected by consolidation (e.g., distractor removal reduces overlap with the pre-consolidation mask).
We evaluate $\tau_{\mathrm{lo}}\in\{0.1,0.3\}$ around the default $\tau_{\mathrm{lo}}=0.2$.
A smaller value ($0.1$) retains more corrected tracker masks, slightly increasing FP, whereas a larger value ($0.3$) prunes more aggressively but may remove some valid corrected tracks. The differences are minor (Table~\ref{tab:ablation_suppl}), confirming that consolidation already effectively suppresses unstable masks.

\paragraph{High overlap threshold $\tau_{\mathrm{hi}}$.}
The high threshold $\tau_{\mathrm{hi}}$ is used to identify \emph{high-agreement} tracker--detector pairs for category-level memory updates and to detect pre/post-con\-so\-li\-da\-tion changes that trigger distractor-removed-driven DRM updates.
We evaluate $\tau_{\mathrm{hi}}\in\{0.75,0.85,0.95\}$ around the default $\tau_{\mathrm{hi}}=0.8$.
Lowering $\tau_{\mathrm{hi}}$ increases the number of candidate updates (more frequent memory refreshes) with a cost of a modest FN increase.  
Increasing $\tau_{\mathrm{hi}}$ to $0.95$ makes updates rare and reduces the opportunity to adapt memory, leading to a small drop in HOTA/MOTA/IDF1 (Table~\ref{tab:ablation_suppl}). The default $\tau_{\mathrm{hi}}=0.8$ provides a conservative update regime.

\paragraph{Overall robustness to thresholds.}
Across all ablations in Table~\ref{tab:ablation_suppl}, varying any single conservative gate leads to only marginal performance changes. The largest observed deviation from the default configuration is observed when we set $\tau_{\mathrm{md}}=0.45$, corresponding to merely \textbf{2\%} decrease in MOTA. Over the same sweeps, HOTA varies by at most less than \textbf{1\%}.
These results demonstrate that the proposed thresholds function as \emph{robust gates} rather than delicately tuned hyperparameters. Performance remains stable across a wide operating range, indicating that \name{} does not depend on dataset-specific calibration or fine-grained threshold optimization.

\section{Application to Single-target Tracking}  \label{sec:st_tracking}

Although \name{} is primarily designed for tracking multiple objects within a specified category, it can be naturally adapted to the single-object tracking (SOT) setting, where only a single instance annotated in the first frame must be tracked throughout the sequence.
To further test \name{}'s cross-task generalization capability, we evaluate it on a recent challenging dataset for single-target tracking DiDi~\cite{dam4sam}.

\begin{table}
\centering
\caption{Single-target tracking performance on the DiDi dataset~\cite{dam4sam} measured using the following standard performance measures: tracking quality, accuracy and robustness.}
\setlength{\tabcolsep}{8pt}
\begin{tabular}{llll}
\toprule
Method & Quality & Accuracy & Robustness \\
\midrule

SAM2.1 & 0.649 & 0.720 & 0.887 \\

DAM4SAM & 0.694 \first{} & 0.727 \first{} & 0.944 \first{} \\

\name{} & 0.685 \second{} & 0.724 \second{} & 0.932 \second{} \\

\bottomrule
\end{tabular}
\label{tab:didi}
\end{table}

Only a minimal modification of the original algorithm is required: we disable instance termination for the initialized target.
This adjustment ensures that the tracker remains active even during extended periods of occlusion or temporary disappearance - scenarios that occur more frequently in SOT than in gMOT - to enable target re-detection.

Table~\ref{tab:didi} compares \name{} with the tracking foundation model SAM2.1~\cite{sam2} and its recent state-of-the-art SOT extension DAM4SAM~\cite{dam4sam}.
\name{} surpasses SAM2.1 by 5.5\% and achieves performance comparable to DAM4SAM, with only 1.3\% lower tracking quality.
Considering that \name{} is fundamentally a multi-target tracker, this result highlights its strong generalization capability across different tracking paradigms.

\section{Qualitative Results}  \label{sec:abl-qualitative}

\begin{figure*}[t]
  \centering
   \includegraphics[width=\linewidth]{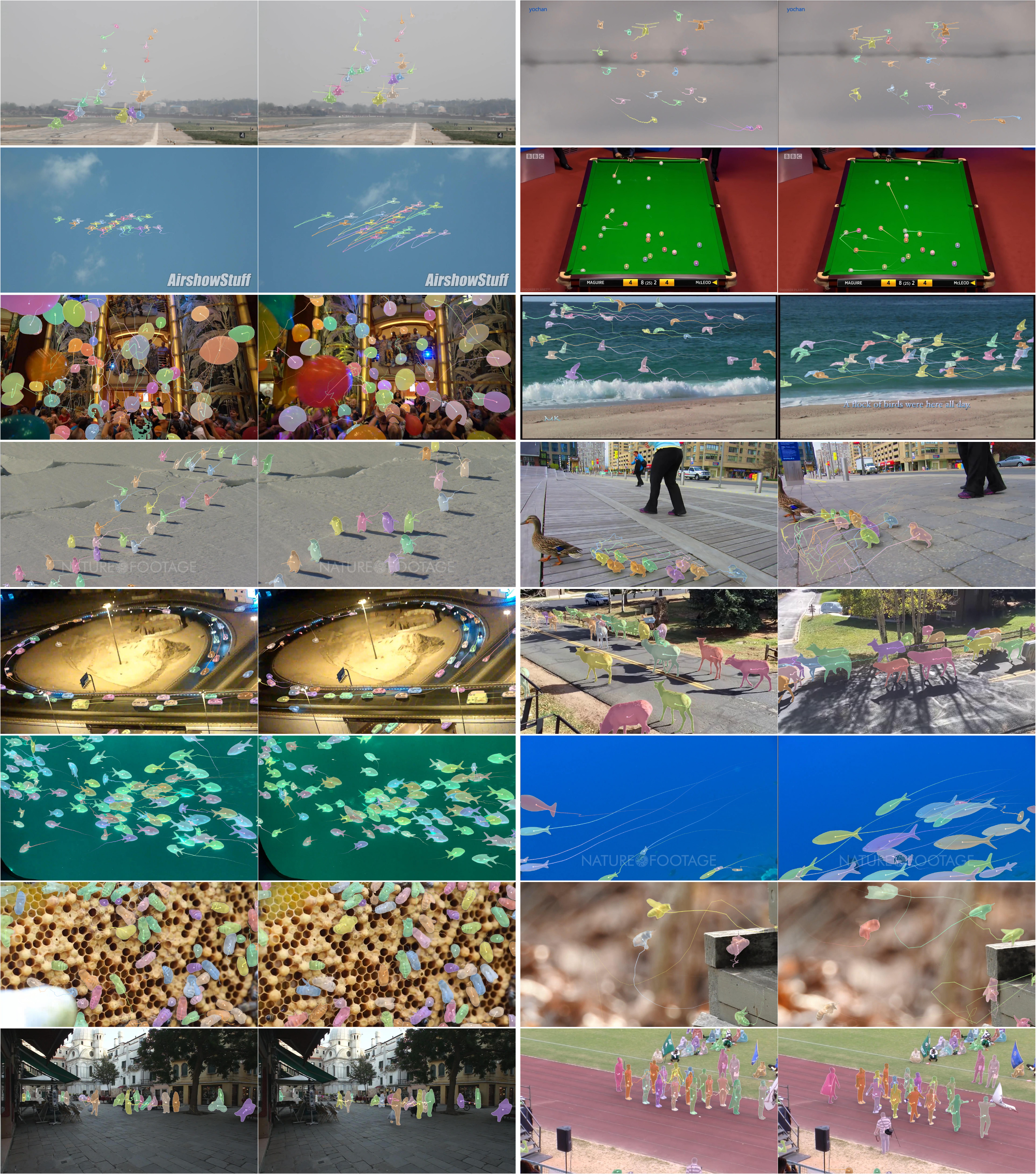}
   \caption{Qualitative examples of tracking with \name{}. Each instance is represented by a colored mask and line, denoting its trajectory.}
   \label{fig:qualitative}
\end{figure*}

Additional qualitative results of multi-object tracking with \name{} are presented in Figure~\ref{fig:qualitative}.
Each tracked instance is visualized using a semi-transparent colored mask, accompanied by a trajectory line indicating its motion over past frames.
We show 16 video sequences from different datasets, displaying two representative frames per sequence.
The examples span a wide range of object categories -- including airplanes, balls, balloons, and various animals (e.g., birds, penguins, ducks, deer, bees, fish), as well as people -- highlighting the strong cross-category generalization capability of \name{}.

\end{document}